\documentclass[letterpaper, 10 pt, journal, twoside]{ieeetran}
\usepackage{times}

\usepackage[numbers]{natbib}
\usepackage{multicol}
\usepackage[bookmarks=true]{hyperref}
\usepackage{amsfonts}
\usepackage{graphicx}
\usepackage{amsmath}
\usepackage{textcomp}
\usepackage{algorithm}
\usepackage{algpseudocode}
\usepackage{booktabs}
\usepackage{multirow}
\usepackage{array}
\usepackage{subcaption}
\usepackage{caption}
\usepackage{authblk}

\begin{document}




\title{Train-Small Deploy-Large: Leveraging Diffusion-Based Multi-Robot Planning}


\author[1]{Siddharth Singh}
\author[2]{Soumee Guha}
\author[1]{Qing Chang}
\author[2]{Scott Acton}

\affil[1]{Department of Mechanical \& Aerospace Engineering, University of Virginia}
\affil[2]{Department of Electrical \& Computer Engineering, University of Virginia}



%

\maketitle

\begin{abstract}
Learning based multi-robot path planning methods struggle to scale or generalize to changes, particularly variations in the number of robots during deployment. Most existing methods are trained on a fixed number of robots and may tolerate a reduced number during testing, but typically fail when the number increases. Additionally, training such methods for a larger number of agents can be both time consuming and computationally expensive. However, analytical methods can struggle to scale computationally or handle dynamic changes in the environment. In this work, we propose to leverage a diffusion model based planner capable of handling dynamically varying number of agents. Our approach is trained on a limited number of agents and generalizes effectively to larger numbers of agents during deployment. Results show that integrating a single shared diffusion model based planner with dedicated inter-agent attention computation and temporal convolution enables a train-small deploy-large paradigm with good accuracy. We validate our method across multiple scenarios and compare the performance with existing multi-agent reinforcement learning techniques and heuristic control based methods.
\end{abstract}

\IEEEpeerreviewmaketitle

\section{Introduction}
Multi-Agent Systems (MAS) play a vital role in modern applications such as warehouse automation, material handling, surveillance, and manufacturing operations~\cite{MASReview}. The problem of multi-agent planning has been pursued for several decades ~\cite{MRS_review},  with a central challenge being efficient path planning and coordination among agents, especially in dynamic environments where the number of agents may vary over time. However, most existing methods assume a fixed number of robots or agents.

Traditional approaches to multi-agent planning can be broadly categorized into two broad classes: analytical or algorithmic methods and learning-based methods. Analytical methods, such as Multi Agent Path Finding (MAPF)~\cite{mapfreview} methods or Nature Inspired methods~\cite{nature_insp}, typically rely on closed-form formulations and offer performance guarantees. However, these methods can struggle to scale due to computational complexity, especially as the number of agents increases, and they can be brittle in dynamic or partially observed environments. Learning-based approaches, such as those based on graph architectures~\cite{graph_learning_survey} and Multi-Agent Reinforcement Learning (MARL)~\cite{marlreview}, have shown greater adaptability in dynamic scenarios. However, their reliance on a fixed number of agents during training constrains their ability to generalize to deployments with larger agent populations. Moreover, these methods suffer from increased non-stationarity and instability in learning when the number of agents changes dynamically~\cite{nonstation}.

Recent advances in generative modeling, particularly diffusion models, offer promising capabilities for handling complex, high-dimensional planning tasks under varying conditions~\cite{genaireview}. Inspired by these advancements, we explore the use of diffusion-based planning for multi-robot systems operating in dynamic and variable-size team settings. Our key goal is to develop a generalizable approach that supports a \textit{train-small, deploy-large} paradigm – training on a limited number of agents while scaling to larger teams at deployment. We leverage the same behavior to handle a system with a dynamically varying number of agents in deployment.

To this end, we investigate the efficacy of diffusion models in improving generalization for multi-agent navigation tasks. Our proposed approach, Multi-Agent Diffusion Based Planner (MA-DBP) begins by leveraging pre-trained MARL policies to bootstrap diffusion model training. This allows the diffusion model to learn a strong prior over agent behaviors, while decoupling planning from the non-stationarity challenges of MARL. Additionally, we design a semantic axial pre-processing module to encode both inter-agent interactions and temporal context, which is critical in enhancing coordination and ensuring trajectory consistency. We adopt a conditional latent diffusion framework, where the diffusion model is conditioned on tokens that encode the current environmental state, the states of active agents, and the target goals. This enables the model to up-sample effective plans even when deployed in configurations with more agents than seen during training. Most importantly, we utilize a single diffusion model to plan and upscale. A moving window planning strategy is employed to maintain goal-directed behavior throughout execution. In summary, our key contributions are as follows:

    
    \begin{enumerate}
    \item A novel diffusion-based planner architecture enabling scalable multi-agent coordination and robust generalization to team sizes larger than seen during training.
    \item  A moving window conditional diffusion framework that flexibly incorporates environment, agent count, and goal conditions to ensure dynamic interaction and scalable goal-reaching behavior in multi-agent setups. 
    \item A semantic axial attention pre-processor that embeds the inter-agent attention and temporal dependencies to support coordinated and coherent planning. All embeddings obtained from this module have the same dimension, which ensures that the model can work with a varying number of agents.
    \end{enumerate}
We validate our approach for the navigation problem with different scenarios in simulation and compare against existing methods. Our results show that leveraging diffusion with explicit attention based encoding prior to noising step can allow us to follow the train-small deploy-large paradigm with good success and and handle dynamic setups.

The rest of the paper is structured as follows; Section~\ref{sec:lit} discusses the related work, Sec.~\ref{sec:prop} details the proposed method, Sec~\ref{sec:exp} shows the experimental validation and comparative analysis against existing methods, followed by the conclusion in Sec~\ref{sec:conclusion}.


\section{Literature Review} \label{sec:lit}
    Research in multi-agent systems can be algorithmic, analytical, and learning-based~\cite{MAS_book}. Algorithmic methods address scheduling and planning tasks like path finding and conflict resolution, using an algorithmic planner with low-level controllers~\cite{mas_algo}. Analytical methods leverage centralized models and optimization frameworks to model objectives and constraints, but face scalability issues as agent numbers grow~\cite{piccoli2023controlmultiagentsystemsresults}. Learning-based methods, including graph learning and MARL, adapt to environmental and task changes, but require retraining for varying agent counts or changing graph topologies due to non-stationarity~\cite{marlreview}.
    
    Recent work on diffusion based planning has shown promising results for utilizing diffusion for generative planning and has been further extended for planning and prediction for multi-agent systems. MotionDiffuser~\cite{jiang2023motiondiffuser} leverages diffusion for trajectory prediction with scenario and agent context. In ~\cite{xu2024sidiff}, diffusion is used for predicting the goal state of the scenario with local observations of each state, however, no planning is involved. In~\cite{shaoul2024multi}, diffusion has been leveraged for path following approach for multi-agent systems, but the planning is off-loaded from the diffusion model. In~\cite{liang2024multi} score-based diffusion learning is used for the path finding problem on a continuous map; however, due to the projection in each diffusion step during optimization, this could lead to high computational cost as the number of agents increases. Similarly, ~\cite{liang2025discreteguideddiffusionscalablesafe} and ~\cite{liang2025simultaneousmultirobotmotionplanning} utilize a constrained optimization directly into the diffusion sampling process, with~\cite{liang2025discreteguideddiffusionscalablesafe} specifically using a projection approach for multi-robot path finding and \cite{liang2025simultaneousmultirobotmotionplanning} focusing on utilizing existing discrete MAPF solvers improving scalability. However, these methods solve the complete offline path finding and have not been validated for dynamic environments or setups.
    
    In~\cite{MizutaLeung2024} and ~\cite{scidiff}, a control-based approach is introduced to provide formal guarantees during denoising handling dynamic obstacles, but is limited to planning for just one robot. MADiff~\cite{zhu2024madiff} is one of the most prominent works on diffusion based planning for MAS, modeling inter-agent interactions using attention between the agents within the diffusion model. However, it is trained with a fixed number of agents limiting the ability to handle a varying number of agents in execution unless an independent denoising network is used for each agent. Similarly, ~\cite{idoko} diffuses the coefficients of a polynomial trajectory for collision avoidance using expert demonstrations. In ~\cite{navformer}, a causal decision transformer is trained on single-robot exploration and multi-robot collision avoidance datasets, which focuses on single-robot target-driven navigation using egocentric visual observations, unlike our multi-agent coordination problem where multiple robots must simultaneously plan collision-free trajectories to different goals.

\section{Methodology}\label{sec:prop}
\subsection{Problem Formulation}
We focus on the problem of 2-D robotic navigation environment with a critical focus on dynamically varying number of agents. The problem is formulated as a goal-conditioned moving horizon navigation problem with a varying number of agents. At any time $t$, given a set of active robots, the current positions $\boldsymbol{x}(t) = [\mathbf{s}_1(t), \mathbf{s}_2(t), \cdots \mathbf{s}_{n_a}(t)]^\top$, and given the final desired goal positions $\mathbf{x}^* = [\mathbf{s}_1^*, \mathbf{s}_2^*, \cdots \mathbf{s}_{n_a}^*]^\top$, the objective is to find the trajectory $\mathcal{T}_{n_a} = [\mathbf{\tau}_1, \mathbf{\tau}_2, \cdots, \mathbf{\tau}_{n_a}]^\top$ for all the agents for the fixed horizon $[t:t_H]$, where $H$ is the length of the horizon and $n_a$ represents the number of active agents for that horizon--drawing attention from real-world scenarios, the number of active agents changes dynamically. 

\begin{figure*}[ht]
    \centering
    \includegraphics[width=0.95\linewidth]{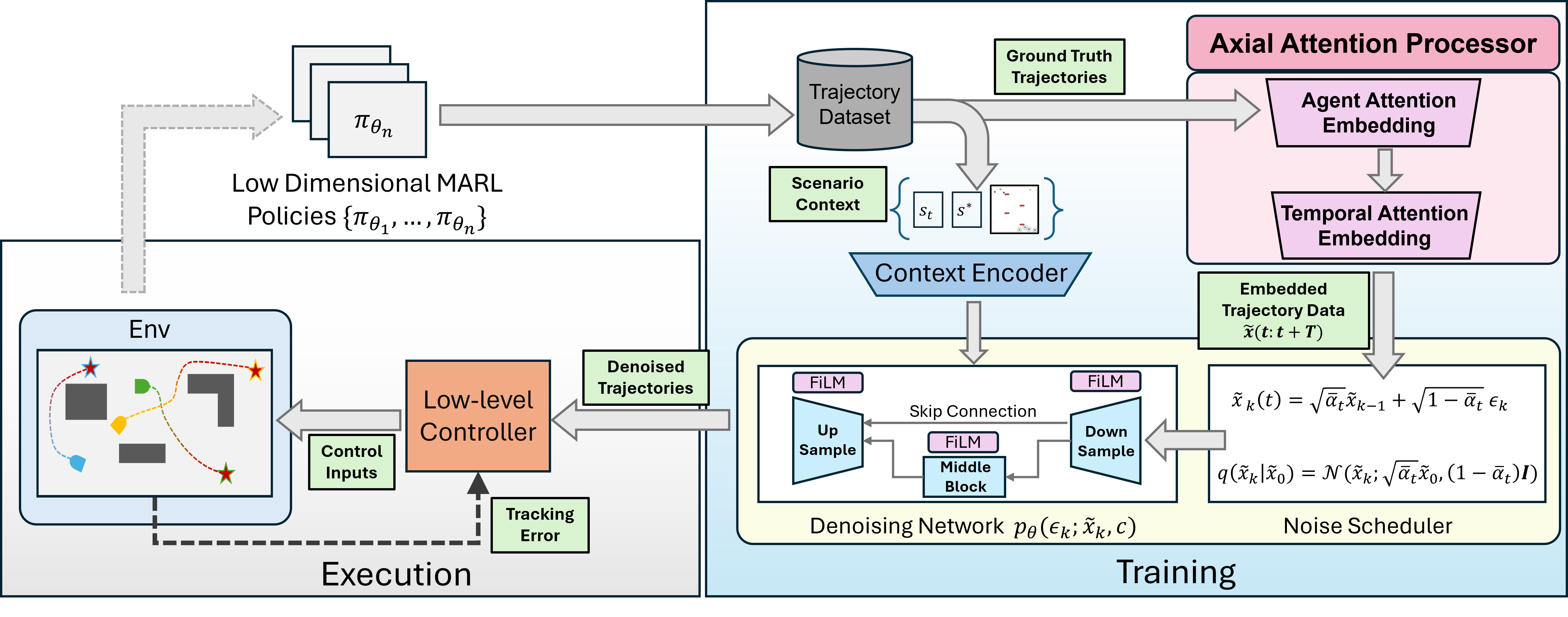}
    \caption{The proposed Multi-Agent Diffusion Based Planner.}
    \label{fig:proposed}
\end{figure*}

\subsection{Proposed Method}


To tackle the challenges for dynamic multi-robotic planning, we adopt a moving window diffusion-based planning approach coupled with a two dimensional semantic axial attention processor and an enhanced network architecture. This emphasizes on abstracting the inter-agent interaction over the length of the trajectory, which is critical to help upscale the method to larger number of agents. The proposed method only uses a single shared model for all the agents and the axial attention processor is the critical component that allows the model to generalize the behavior to larger number of agents. To enable a goal-conditioned planning method and to bolster the ability to handle dynamically varying number of agents, a conditional diffusion approach has been incorporated by including context encoding. 

Figure \ref{fig:proposed} shows the schematic diagram of the proposed Multi Agent Diffusion Based Planner (MA-DBP). We first leverage existing MARL methods to collect a dataset of trajectories trained with smaller number of agents. We then sample trajectories from this dataset to train a goal conditioned diffusion planner. Additionally, we also collect scenario specific data for generating context tokens for the denoising network. The trajectory data is pre-processed by passing it through the Axial Attention Pre-processor, and after a noise scheduler is used to generate the noisy signal, a U-Net based denoising procedure is employed to learn the added noise. Additional losses are introduced to enforce temporal consistency, maintain feasible boundary conditions and reduce collisions. The encoded context information is provided at each step of U-Net with attention modules. Finally, during execution, the denoised trajectory is passed to a low-level controller, ensuring that the trajectory is executed with safety and minimizing local collisions. Decoupling the low-level controller from the planner allows for training the method for heterogeneous systems. We provide further details of our method in the following subsections.

\subsection{Dynamic Multi Robot Path Planner based on Enhanced Diffusion Method}
The fundamental principles of the proposed method aims to learn the low-dimensional policies and utilize that to upscale to higher dimensions. This implies an aggregated learning approach. The proposed MA-DBP leverages three main components to do so; firstly, by training with distinct number of agents in a moving window fashion, secondly, by carefully embedding the trajectory data to a higher dimension capturing the multi-agent behavior leveraging attention, and lastly, by constraining the trajectories to maintain feasibility with a multi-component loss function.


\subsubsection{Moving Window Diffusion with Enhanced Attention Processor}

To ensure that we can capture the dynamic behavior, we pose the problem as a moving window trajectory planning problem with a fixed window horizon. It is also important to notice that the length of the horizon is a key design parameter. If the planning horizon is too short, the model may fail to converge due to limited temporal context and feature sparsity. On the other hand, if the horizon is too long, the moving window may capture overly static patterns and overlook important dynamic transitions, making it difficult to generate high-quality, goal-directed trajectories. Moreover, the increased dimensionality of long-horizon planning complicates training and can lead to less stable or less focused behavior. To address this, we sweep across different horizon lengths to identify an optimal window size that balances dynamic awareness, training stability, and computational tractability.

\subsubsection{Attention Processing and Network Architecture}
\textbf{Axial Attention Processor}:
Unlike existing methods~\cite{zhu2024madiff, jiang2023motiondiffuser} which include the attention within the diffusion model, we explicitly embed the trajectory with different semantic dimensional attention processing block before the denoising step. Video diffusion~\cite{video_diffusion} based methods that have temporal attention embedding in the U-Net architecture, struggle to diffuse with such low-dimensional trajectory data and cannot handle task constraints. Transformer based methods, which again have attention blocks within the denoising network, are severely data hungry and significantly hard to train~\cite{DIT_diff_transformer}. The proposed model learns the inter-agent behavior through this module and can thus generalize for a varying number of agents, without additional supervision. 

To be able to generalize the behavior from a smaller number of agents to a larger number of agents, it is critical to capture two things, i) the inter-agent interaction which can include things such as collision avoidance, and ii) the temporal evolution of the trajectories over time. Assuming that we have the trajectory data for a batch size of $B$ trajectories, $X^k \in \mathbb{R}^{B \times H \times n_a \times d}$, where $d$ is the dimension of state. The input is first projected to a higher embedding space $X^k_{proj} = Linear(X^k) \in \mathbb{R}^{B \times H \times n_a \times D}$. Combining with the learned positional embedding for the number of agent dimensions ($PE_{Agent}$) and temporal dimensions ($PE_{temp}$), resulting in the positional embedding variable,
\begin{equation}
    X^k_{pos} = X^k_{proj} ~+ PE_{Agent}(X^k) ~+ PE_{temp}(X^k) 
\end{equation}
where $X^k_{pos}$ are the learned positional embedding function that maps the input variable to the same high-dimension as the projection space, $PE : \mathbb{R}^{B \times H \times n_a \times d} \rightarrow \mathbb{R}^{B \times H \times n_a \times D}$.
The positional embedded variable is first transformed to $X_{agent}^k = \text{Reshape}(X_{pos}^k) \in \mathbb{R}^{(B \times n_a) \times H \times D}$. Using this, we compute the attention embedding along the number of agent axis, resulting in $A_{agent} = Attention(X^k_{agent}) \in \mathbb{R}^{B \times H \times n_a \times D}$. After a layer-norm is applied this results in $X^k_{a} = \text{LayerNorm}(X^k_{pos} + A_{agent})$. Next we compute the attention along the time axis, by first transforming the embedded variable to $X^k_{time} = \text{Reshape}(X^k_{a}) \in \mathbb{R}^{(B \times H) \times n_a \times D}$. After computing the attention along the time axis this results in $A_{time} = Attention(X^k_{time})\in \mathbb{R}^{B \times H \times n_a \times D}$. Finally after combining it with the positional embedded variable and parsing through a layer-norm, we get $X^k_{at} = LayerNorm(X^k_{pos} ~+ A_{time}) \in \mathbb{R}^{B \times H \times n_a \times D}$. Finally, we pass the attended features through a MLP for feature refinement resulting in $\tilde{X}^k = X^k_{at} + MLP(X^k_{at}) \in \mathbb{R}^{B \times H \times n_a \times D}$ resulting in the attention embedded variable in higher dimension. 

Unlike existing methods which leverage inter agent interaction in the context and a single diffusion model for each agent, the proposed axial-attention processor block embeds this before passing it to the U-Net. This reduces the need to repeat the diffusion block for each agent and minimize the parameters of the model. 

\textbf{Network Architecture and Context Encoding}: The network architecture is designed to accommodate larger than $n^{train}_{max}$ agents through masking mechanisms that ensure seamless training and execution with variable agent numbers, supporting both complete trajectory generation and moving horizon training via careful data chunking. During training, the noise-free signal $X^0(t:t_H)$ undergoes $k$ diffusion steps to produce noisy signal $X^k(t:t_H)$, which is then processed through the Axial Attention Processor to yield the embedded diffused variable $\bar{X}^k(t:t_H) \in \mathbb{R}^{B \times H \times n_a \times D}$. The denoising network follows a U-Net encoder-decoder architecture with skip connections, comprising three essential blocks: the downsampling block (two-layer 1D convolution with batch normalization, ReLU activation, and max-pooling), the middle block (similar convolution architecture without pooling to maintain dimensional consistency), and the upsampling block (two-layer 1D convolution with ReLU activation and batch normalization after skip-connection concatenation), culminating in a decoder block that transforms the predicted noise back to the original trajectory dimensions $\epsilon_{\theta}(t:t_H) \in \mathbb{R}^{B \times H \times n_a \times d}$.

Since the aim of the diffusion model is to generate a goal-conditioned multi-agent planner, we provide the scenario image frame $O_t$, initial agent positions $\mathbf{x}_t$, goal positions $\mathbf{x}^*$, and agent count $n_a$ as context information to the denoising network. The unprocessed context token is defined as $c = \{O_t, \mathbf{x}_t, \mathbf{x}^*, n_a\}$ built using the information captured as discussed earlier. We generate embedded context representations through a multi-modal encoder comprising: (1) a CNN-based image encoder for scenario understanding, (2) an MLP-based pose encoder for start/goal positions, and (3) learned embeddings for agent count. These modalities are fused using multi-head attention to produce the final embedded context $\tilde{\mathbf{c}} = \text{Attention}\left(\text{CNN}(O_t), \text{MLP}([\mathbf{x}_t, \mathbf{x}^*]), \text{Embed}(n_a)\right)$. The context is further augmented with sinusoidal time embeddings and integrated into each U-Net level through Feature-wise Linear Modulation (FiLM)~\cite{film}, enabling adaptive conditioning based on both spatial context and diffusion timestep.

\subsubsection{Losses}
To ensure that the sampled trajectories are dynamically feasible and have temporal consistency, we use a multi-term loss to include the loss based on boundary constraints and to minimize jerk, as well as maintain a temporally feasible trajectory. 
\begin{equation}\label{eq:diff_loss}
  \mathcal{L}_{\text{noise}} = \mathbb{E}_{k,\epsilon \sim \mathcal{N}(0,I)} \left[ \|\epsilon - \epsilon_\theta(\mathbf{X}^k, k, \mathbf{c})\|_2^2 \right]  
\end{equation}
Eq. \ref{eq:diff_loss} represents the standard diffusion loss, training the network to learn the added noise. As highlighted in~\cite{DDPM}, minimizing the loss based on noise prediction or the initial signal is equivalent. This allows us to combine the noise loss with the other losses that control the behavior of the predicted trajectory. 
\begin{equation}\label{eq:bond_loss}
    \mathcal{L}_{\text{boundary}} = \frac{1}{B} \sum_{b=1}^{B} \left[ \mathcal{L}_{\text{start}}^{(b)} + \mathcal{L}_{\text{goal}}^{(b)} \right]
\end{equation}
Eq~\ref{eq:bond_loss} is introduced to ensure that the initial and final poses of the denoised trajectory match the requirements. Here $\mathcal{L}_{\text{start}}^{(b)} = ||\hat{\mathbf{x}}^0(t)  - \mathbf{x}(t)||_2$ and $\mathcal{L}_{\text{goal}}^{(b)} = ||\hat{\mathbf{x}}^0(t_H)  - \mathbf{x}^{*}||_2$. It should be noted that these losses are computed after  denoising for each horizon and $x^*$ is the final desired goal and a moving horizon goal. This bolsters the diffusion planners goal-seeking behavior.

\begin{equation}\label{eq:jerk_loss}
    \mathcal{L}_{\text{temporal}} = \frac{1}{B} \sum_{b=1}^{B} \left[ \mathcal{L}_{\text{direction}}^{(b)} + \lambda_{\text{jerk}} \mathcal{L}_{\text{jerk}}^{(b)} \right]
\end{equation}
Additionally, we include a loss which forces the diffusion network to generate temporally consistent trajectories, and minimize jerk as shown in eq.~\eqref{eq:jerk_loss}. Where 
\begin{equation}
  \mathcal{L}_{\text{acc}}^{(b)} = \frac{1}{n_a} \sum_{i=1}^{n_a} \|\mathbf{v}_{b,i,t+1} - \mathbf{v}_{b,i,t}\|_2 ~~
 \end{equation}
and
 \begin{equation}
 \mathcal{L}_{\text{jerk}}^{(b)} = \frac{1}{n_a} \sum_{i=1}^{n_a} \|\mathbf{a}_{b,i,t+1} - \mathbf{a}_{b,i,t}\|_2  
\end{equation}
Here $\mathbf{a}$ and $\mathbf{v}$ are the acceleration and velocity of the trajectory respectively, computed by finite differences with zero-order hold. 
Lastly, we also include the loss based on collision, which is defined by
\begin{equation}
\begin{split}
\mathcal{L}_{\text{collision}} = \frac{1}{B} \sum_{b=1}^{B} \sum_{t=1}^{T} \sum_{i,j \in \mathcal{A}_b, i \neq j} \text{ReLU}(d_{\min} - \|\mathbf{s}_{b,i,t} - \mathbf{s}_{b,j,t}\|_2) + \\ \sum_{i \in \mathcal{A}_b} \sum_{o \in \mathcal{O}_b} \text{ReLU}(d_{\text{obs}} - \|\mathbf{s}_{b,i,t} - \mathbf{o}\|_2) 
\end{split}
\end{equation}

Here, $\mathbf{o}$ is the position of the obstacle center position. 

This results in the final loss 

\begin{equation}
    \mathcal{L}_{\theta} = 
    W^T[\mathcal{L}_{\text{noise}};
    \mathcal{L}_{\text{boundary}}; 
    \mathcal{L}_{\text{temporal}}; 
    \mathcal{L}_{\text{collision}}
    ]
\end{equation}

where $W$ is a weighting vector allowing to tune the different loss components. Our experiments highlight that giving maximum weight to the $\mathcal{L}_{\text{noise}}$ component was important for fast and stable training. The weight vector used for this case is $W = [0.85, 0.025, 0.025, 0.1]^\top$. 

\subsection{Training and Execution}\label{sec:train}
\textbf{Data Collection}
To train the diffusion model, we leverage MARL techniques to collect data. We train multiple policies, $\pi_{\theta_n}$, where $n \in \{1, \cdots, n^{train}_{max}\}$, where $n^{train}_{max}$ is the maximum number of agents in the scenario during training. The starting poses and final goals are sampled randomly for each agent. For collecting the trajectory dataset, we store the position and velocity of each agent along with the start $\mathbf{s}_t$ and final goal positions $\mathbf{s}^*$. Additionally, we also store the actions (control inputs).
For the purpose of context tokens, we store frame for the initial point in the trajectory $O_t$ depicting the scenario as an RGB image frame. Given a horizon $H$, for a $n$ agents we sample a trajectory $x_n (t:t_H) \leftarrow \mathcal{T}\left(\pi_{\theta_n}(s_t,s*)\right)$, where $\mathcal{T}(\pi(s_i,s_f))$ is a function which stores the trajectory data given a policy $\pi_{\theta_n}$, the initial state $s_i$ and the final state $s_f$. For simplicity, throughout the rest of this section we store agent positions in the trajectory. 

Given the noise free embedded input $\tilde{x}^0$ at step $k = 0$, the \textit{forward process} or the \textit{diffusion process} focuses on generating the approximate posterior $q(\tilde{x}^{1:K}|x^0)$ with a fixed Markov chain given as
\begin{equation}
\begin{split}
    q\left(\tilde{x}^{1:K}|\tilde{x}^0\right) := \prod^K_{k=1} q\left(\tilde{x}^k|\tilde{x}^0\right) , \\ q\left(\tilde{x}^k|\tilde{x}^{k-1}\right) := \mathcal{N}\left(\tilde{x}_k; \sqrt{1-\beta_k}\tilde{x}^k, \beta_k \mathbf{I})\right)
\end{split}
\end{equation}
Here $\beta_k$ is the noise variance at diffusion step $k$, $\mathbf{I}$ is the identity matrix and $\mathcal{N}(\cdot)$ represents the normal distribution.


Following this, the second Markov Chain is utilized to learn a parametrized joint distribution, $ p_{\theta}\left(\tilde{x}^{K:0}\right)$. The diffusion model can be further extended to include the condition based diffusion process, where given the condition $c$, the denoising sampling is computed as 
\begin{equation}
\begin{split}
    p_{\theta}\left(\tilde{x}^{K:0}\right):= p_\theta(\tilde{x}^K)\prod^K_{k=1}p_\theta \left(\tilde{x}^{k-1}|\tilde{x}^k;c\right), \\ 
    p_\theta(\tilde{x}^{k-1} | \tilde{x}^k, c) = \mathcal{N}(\tilde{x}^{k-1}; \mu_\theta(\tilde{x}^k, k, c), \Sigma_\theta(\tilde{x}^k, k, c))
    \end{split}
\end{equation}


\textbf{Training Procedure}
To ensure that the proposed approach can handle varying number of agents we employ two steps: i) a curriculum learning based approach and ii) a agent masking approach in training. The number of agents in training varies from $n_{train} = \{1,\cdots, n^{max}_{train}]\}$. The curriculum approach is utilized to ensure a stable learning procedure in which the number of agents increases linearly every $1000*n^{max}_{train}$ epoch. To ensure that the model can handle a larger number of agents in execution, the network is defined for $n_{max}$ agents, and masking is applied to identify only the active agents. It should be noted that $n^{max}_{train} < n_{max} $.

\begin{algorithm}
\caption{Enhanced Multi-Agent Diffusion Model Training with Curriculum Learning}
\label{alg:madp_training}
\begin{algorithmic}[1]
\Require Dataset $\mathcal{D}$, max agents $n_{\max}^{\text{train}}$, epochs $E$, horizon size $H$
\Require Model $\epsilon_\theta$, optimizer, learning rate $\eta$, batch size $B$
\Ensure Trained diffusion model $\epsilon_\theta^*$

\State Initialize model parameters $\theta$
\State Initialize optimizer with learning rate $\eta = 3 \times 10^{-4}$

\For{epoch $e = 1$ to $E$}
    \State $n_{\text{curr}} \gets \text{CurriculumSchedule}(e, E, n_{\max}^{\text{train}})$
    \State $\text{progress} \gets e / E$
    
    \For{batch $\{\mathbf{F}, \mathbf{S}, \mathbf{G}, \mathbf{N}, \mathbf{T}\} \in \mathcal{D}$}
        \State Filter batch to agents $\leq n_{\text{curr}}$
        
        
        \State Update $\theta$ using $\nabla_\theta \mathcal{L}_{\text{batch}}$
        \State Clip gradients: $\|\nabla_\theta\| \leq 1.0$
    \EndFor
    
\EndFor

\State \Return $\epsilon_\theta^*$
\end{algorithmic}
\end{algorithm}

\textbf{Execution}
With the proposed architecture, we utilize the context token to define the planning problem. During sampling, we define the current poses of each agent, the desired goal positions, and the initial scenario. Algorithm \ref{alg:diffusion_sampling} shows the sampling step utilized to compute the path plan for all the active agents. 
\begin{algorithm}
\caption{Diffusion Model Sampling with Boundary Constraints}
\label{alg:diffusion_sampling}
\begin{algorithmic}[1]
\Require  Trained UNet model $\epsilon_\theta$, noise schedule $\{\bar{\alpha}_t\}_{t=1}^T$, conditioning $c$, constraint mask $M$, constraint values $x_{\text{constraint}}$
\Ensure  Generated sample $X_0$

\State  Initialize $\tilde{X}^k \sim \mathcal{N}(0, I)$

\For{$t = T-1, T-2, \ldots, 1$}
    \State  $\epsilon_\theta \gets \text{UNet}(t, c)$
    \State  $\hat{x}_0 \gets \frac{x_t - \sqrt{1 - \bar{\alpha}_t} \cdot \epsilon_\theta}{\sqrt{\bar{\alpha}_t}}$
    \State  $\tilde{x}_{t-1} \gets \sqrt{\bar{\alpha}_{t-1}} \cdot \hat{x}_0 + \sqrt{1 - \bar{\alpha}_{t-1}} \cdot \epsilon_\theta$
    \State  $\tilde{x}_{t-1}[M] \gets \tilde{x}_{\text{constraint}}$ \Comment {Apply boundary constraints}
\EndFor

\Return $\tilde{x}_0$ 
\end{algorithmic}
\end{algorithm}


\section{Experimental Results}
\label{sec:exp}
We utilize the Vectorized Multi-Agent Scenarios (VMAS)~\cite{vmas} framework to train and validate our method. To demonstrate the effectiveness and advantages of the proposed MA-DBP, we compare it with the MARL algorithm, namely MAPPO, a heuristic Lyapunov Function Quadratic Product (CLF-QP) based controller, and a conventional diffusion-based method MADiff   ~\cite{zhu2024madiff}. We leverage pre-trained MAPPO policy to bootstrap training of both diffusion-based models, MA-DBP and MADiff ~\cite{zhu2024madiff}. The training and validation were conducted on a Lambda workstation with AMD Ryzen Threadripper 3970x 32-core processor, 256 GB of RAM with a RTX 3090 GPU and 24 GB of VRAM.
\begin{figure}
    \centering
    \includegraphics[width=1\linewidth]{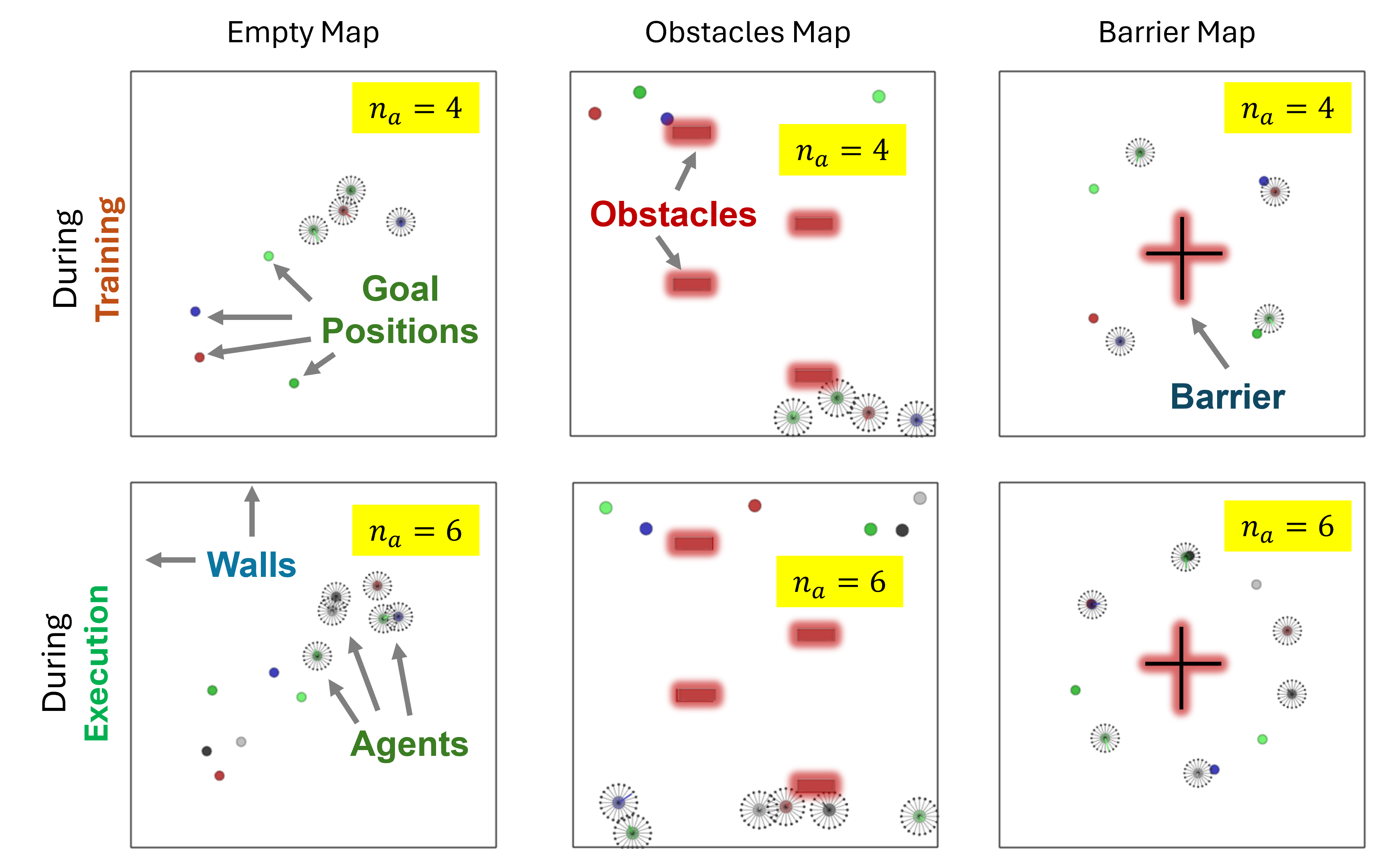}
    \caption{Three navigation scenarios used for validation. (Left to right) Empty map, obstacle map and barrier map.}
    \label{fig:valscenarios}
\end{figure}
All the methods are validated on three 2-D navigation scenarios as shown in Fig.~\ref{fig:valscenarios}, namely navigation in Empty map, Obstacle map and Barrier map. For all the scenarios, the agents are initialized randomly in the map and must navigate to their respective goals which are represented as circles with same colors as agents. Each experiment is repeated 20 times and a successful goal reaching is defined as when each agent is within 0.1 units from it's respective goal position. At the end of the test episodes, we define the average success rate as the ratio of test episodes in which all agents achieve their respective goals successfully within the allowed maximum number of steps. Each simulation episode ran for maximum 100 steps.

Our experiments highlight that the proposed method can upscale in execution with fairly good performance, while also requiring lesser training time. The ablation studies highlight the benefits of utilizing the moving window approach along with the benefits of the Axial Attention Processing.

\subsection{Training and Execution Comparison}
\begin{figure}
    \centering
    \includegraphics[width=\linewidth]{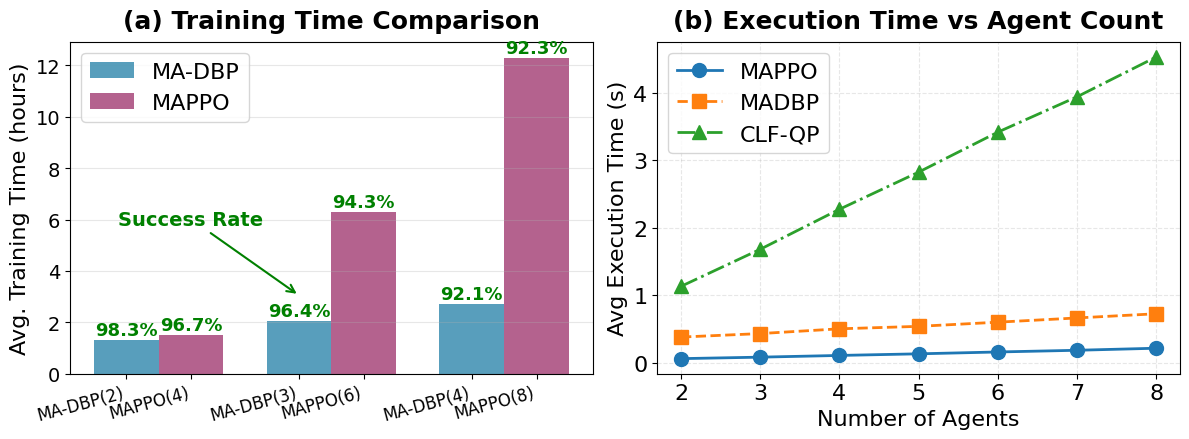}
    \caption{(a) Comparison in training time and success rate for MAPPO($n_a$) and MA-DBP($n_a/2$) agents. The success rate is the success rate when executed on $n_a$ agents. (b) Plot comparing real-time execution comparison between MAPPO, MA-DBP (Proposed) and CLF-QP.}
    \label{fig:train_comp}
\end{figure}

We compare MAPPO and MA-DBP in terms of training efficiency on the empty map (see Fig.~\ref{fig:train_comp}(a)). MAPPO is trained with $n_a$ agents, while MA-DBP is trained using only $n_a/2$ agents, yet both are evaluated on $n_a$ agents. Results show that MA-DBP achieves a comparable success rate while requiring up to 4x less total training time for $n_a^{train}=8$, demonstrating the benefits of the train-small, deploy-large approach. Notably, MA-DBP’s reported time includes both MAPPO pre-training (for data generation) and its own training phase. This highlights MA-DBP’s scalability: efficient learning with fewer agents translates into significant computational savings without sacrificing performance.


Table~\ref{tab:comp_per} compares MA-DBP against MAPPO, CLF-QP, and MADiff~\cite{zhu2024madiff} across three scenarios, with all methods trained and evaluated on identical agent counts. MA-DBP demonstrates competitive performance overall. CLF-QP struggles in cluttered environments due to lack of collision foresight, often getting stuck. MADiff, using classifier-free guidance without explicit goal conditioning, exhibits weaker goal-reaching behavior and struggles with obstacles in our tasks, even though it replans a full horizon‑H trajectory at each time step and executes only the first action. MAPPO, which provides training data for both diffusion methods, serves as the baseline and generally outperforms MA-DBP and MADiff, particularly in obstacle-rich scenarios owing to the fundamental nature of bootstrapping and due to one-step sampling nature of MAPPO. However, MA-DBP's moving horizon approach enables performance comparable to MAPPO in many cases and superior to CLF-QP and MADiff.

\begin{table}[h!]
\centering
\scriptsize
\caption{Performance Comparison (Avg. Success Rate)}
\label{tab:comp_per}
\begin{tabular}{lcccc}
\toprule
 & \textbf{MAPPO} & \textbf{CLF-QP} & \textbf{MADiff} & \textbf{MA-DBP}* \\
\midrule
\midrule
\multicolumn{5}{c}{\textbf{Navigation Empty Map}} \\
\midrule
n=2 & 0.96 $\pm$ 0.024 & \textbf{0.97 $\pm$ 0.020} & 0.55 $\pm$ 0.081 & 0.94 $\pm$ 0.027\\
n=4 & 0.94 $\pm$ 0.033 & 0.96 $\pm$ 0.021 & 0.47 $\pm$ 0.011 & \textbf{0.96 $\pm$ 0.034}\\
n=6 & 0.94 $\pm$ 0.021 & 0.94 $\pm$ 0.036 & 0.42 $\pm$ 0.031 & \textbf{0.95 $\pm$ 0.021} \\
\addlinespace
\midrule
\multicolumn{5}{c}{\textbf{Navigation with Obstacles}} \\
\midrule
n=2 & \textbf{0.94 $\pm$ 0.021} & 0.80 $\pm$ 0.01 & 0.12 $\pm$ 0.021 & 0.72 $\pm$ 0.100\\
n=3 & \textbf{0.95 $\pm$ 0.037} & 0.68 $\pm$ 0.02 & 0.15 $\pm$ 0.018 & 0.70 $\pm$ 0.080\\
n=4 & \textbf{0.94 $\pm$ 0.014} & 0.65 $\pm$ 0.02 & 0.18 $\pm$ 0.010 & 0.78 $\pm$ 0.065\\
\addlinespace
\midrule
\multicolumn{5}{c}{\textbf{Navigation with Barrier}} \\
\midrule
n=2 & \textbf{0.92 $\pm$ 0.035} & 0.78 $\pm$ 0.021 & 0.37 $\pm$ 0.026 & 0.81 $\pm$ 0.048 \\
n=3 & \textbf{0.88 $\pm$ 0.021} & 0.65 $\pm$ 0.016 & 0.32 $\pm$ 0.034 & 0.73 $\pm$ 0.096 \\
n=4 & \textbf{0.84 $\pm$ 0.034} & 0.67 $\pm$ 0.014 & 0.30 $\pm$ 0.041 & 0.65 $\pm$ 0.047 \\
\bottomrule
\end{tabular}

\end{table}

\subsection{Execution Speed Comparison}
We also compare the efficacy of our method in real-time execution speed. Fig.~\ref{fig:train_comp}(b)) shows the plot comparing the average execution time for the empty navigation scenario for the three methods against the number of agents. It was evident that the MAPPO method was the fastest during inference, while the MA-DBP is slower than MAPPO, it still shows good comparative performance. The CLF-QP method sees an increase in execution time with number of agents owing to the increased computational cost. This further highlights the real-world applicability of the proposed MA-DBP. With  an average sampling time of 0.04 seconds for the cases, the proposed methods seems to be able to handle real-time performance. 




\subsection{Quantitative Analysis of Scalability}
Most importantly, we also investigate the ability of the proposed MA-DBP to scale to larger number of agents. We again validate this on three scenarios, an empty map, barrier map and obstacle map. All agents and their respective goals are randomly spawned. 

\begin{figure*}[htbp]
    \centering
    \includegraphics[width=0.95\linewidth]{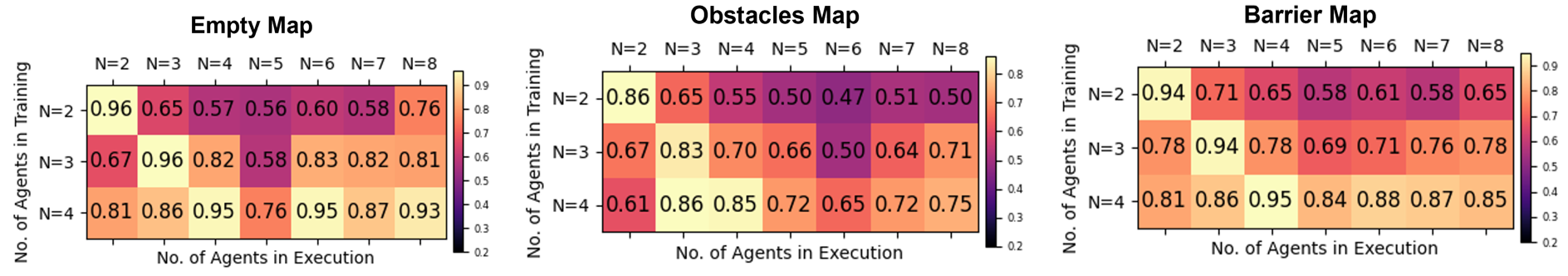}
    \caption{Success rate in upscaling with MA-DBP for three different scenarios with different number of agents in training and execution.}
    \label{fig:upscale}
\end{figure*}

Figure ~\ref{fig:upscale} shows the results from upscaling with proposed MA-DBP approach. As observed from the experiments, the proposed MA-DBP scales with good accuracy even as the number of agents increase. We train the model with smaller number of agents and increase the number of agents gradually during execution. While $n_a = 2$, leads to reasonable performance when upscaling, with $n_a = 3, 4$ the upscaling works significantly better. For $n_a = 2$ the success rate does not scale very well when the number of agents increases, and it appears that this is mostly due to the lack of information to capture enough interactions between agents. Another observation made from the experiments was that the decrease in performance for the larger number of agents was not substantial when $n_a$ increased, especially for $n_a^{train} > 2$. This was attributed to a higher density of goal positions which makes it easier for MA-DBP to plan a path, however, it was also observed that a higher density naturally increased the number of collisions. 
\vspace{-0.12in}

\subsection{Ablation Studies}\label{sec:ablation}
\begin{figure}
    \centering
    \includegraphics[width=0.95\linewidth]{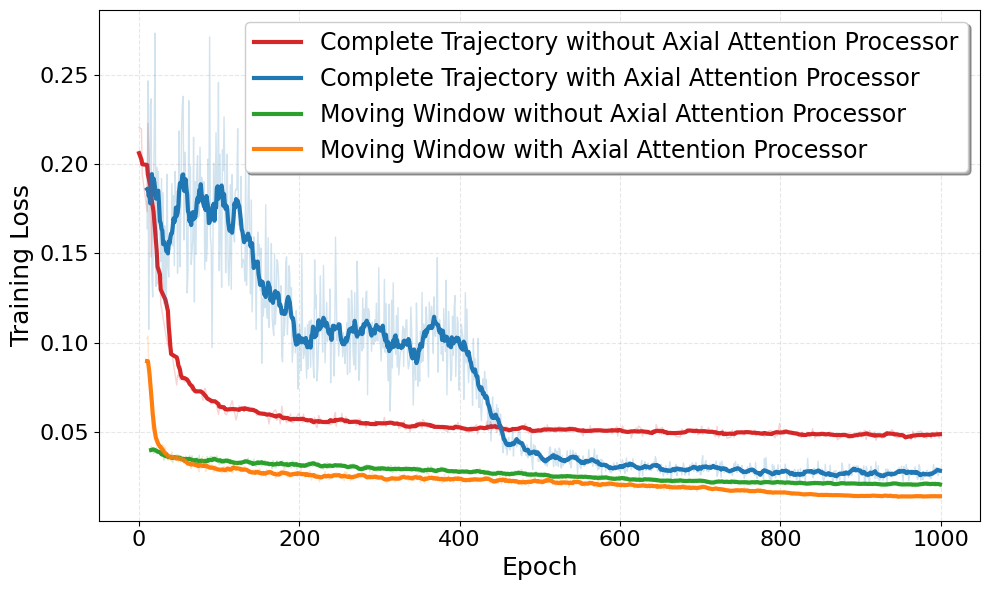}
    \caption{Plot comparing the training loss for the models for ablation study identifying the effect of moving window approach and semantic-axial attention pre-processing.}
    \label{fig:abalation}
\end{figure}

For further highlighting the benefits of the semantic-axial attention processing and the moving window approach, we carry out ablation studies. In the case where the axial attention processor is removed, it is replaced by a linear encoder (LE) which embeds the trajectory to the same dimension as the axial attention processor to ensure a feasible comparison. We compare the training and execution performance of four models, i) complete trajectory without axial attention processor (CT+LE), ii) complete trajectory with axial attention processor (CT+AAP), iii) the model with moving horizon but without axial attention processor (MW+LE) and iv) the proposed axial attention processor with the moving window (MA-DBP). The horizon length for moving window is 10 steps. We train and validate the models for the Empty navigation map. Figure~\ref{fig:abalation} shows the training loss of the four models for $n_a^{train} = 4$. As seen, during training, the complete trajectory model is the slowest in learning. Table~\ref{tab:abal_comp} shows the average success rate in upscaling for the four methods. The first column shows the number of agents in training and the second row shows the number of agents in execution. As is evident, the complete trajectory method often struggles to get near the goal region. Additionally, it also evident that the proposed Axial Attention processor is much more critical in improving the upscaling performance. 

\begin{table}[htbp]
\centering
\caption{Ablation Comparison (Avg. Success Rate)}
\label{tab:abal_comp}
\renewcommand{\arraystretch}{1.2} 
\begin{tabular}{c|c c|c c|c c|c c}
\hline
& \multicolumn{2}{c}{CT + LE}&\multicolumn{2}{c}{CT + AAP} & \multicolumn{2}{c}{MW + LE} & \multicolumn{2}{c}{MA-DBP*} \\
\cline{2-9}
& n=6 & n=7 & n=6 & n=7 & n=6 & n=7 & n=6 & n=7 \\
\hline
n=2 & 0.18 & 0.10 & 0.32 & 0.34 & 0.46 & 0.42 & \textbf{0.60} & \textbf{0.58} \\
n=3 & 0.28 & 0.22 & 0.68 & 0.59 & 0.65 & 0.61 & \textbf{0.83} & \textbf{0.82} \\
n=4 & 0.34 & 0.31 & 0.71 & 0.69 & 0.72 & 0.67 & \textbf{0.95} & \textbf{0.87} \\
\hline
\end{tabular}
\end{table}

\subsection{Discussion}
The proposed MA-DBP method shows promising performance in leveraging diffusion for training a multi-agent planner which can generalize to system settings. Although the experiments highlight different characteristics of the proposed method, we observed different contributing factors. First, the horizon size needs to be chosen carefully. If the horizon is too long the model struggles to capture the nuances of the trajectory. And if the horizon length is too small, the attention module fails to capture inter-agent interaction effectively. The MAPPO outperforms the MA-DBP in cluttered environments, since it can sample one step at a time, whereas the diffusion model struggles to learn when the planning horizon is too short. Secondly, two of the main contributing factors for improving upscalability are the Axial Attention Processor and utilizing a curriculum approach in training. Although the training loss plot in Sec.~\ref{sec:ablation} does not make it apparent enough due to the scale of the losses, the reason the losses do not reduce any further for case 1 (red) in Fig.~\ref{fig:abalation} are because as the number of agents is increasing in a curriculum fashion the model weights fail to adjust and struggle to overcome this issue. Also, while the upscaling success rate is high as the number of agents increase it lead to a higher number of collisions. The test was limited to 8 agents due to the size of the environment. Our experiments highlight that leveraging the explicit attention computation and combining it with curriculum based training and a moving horizon approach is necessary for ensuring upscaling  behavior. Removing either of these three components leads to failure.

\section{Conclusion}
\label{sec:conclusion}

This work investigated a diffusion based planner for multi-agent systems and specifically dived deeper into studying the ability of the diffusion to upscale to larger number of agents. We introduced an axial-attention based approach to capture both inter-agent interaction and temporal composition of the trajectory. By introducing losses on collision avoidance and temporal consistency we were able to devise a conditional diffusion based planner, which generates motion plans for multi-agent robotic systems. Our experiments highlight that even though the method is trained for a smaller number of agents, it can handle, previously unseen, a larger number of agents with acceptable success. This approach highlights the ability of diffusion to generalize the behavior from low-dimensional policies to high-dimensional policies. Additionally, we also highlight that MA-DBP can also outperform existing MARL and control based methods in execution in certain aspects; such as handling dynamically changing agents, sampling speed in execution, and handling larger number of agents.
However, as a implicit behavior with diffusion models, one can either train them for coarse long horizon trajectories, where the goal is far or for fine trajectories where the final goal is within the vicinity. Combining the two behaviors can lead to unstable training. 




\begingroup
\small
\bibliographystyle{plainnat}
\bibliography{references}
\endgroup

\end{document}